\documentclass[letterpaper, 10 pt, conference]{ieeeconf}  

\usepackage{cite}
\usepackage{amsmath,amssymb,amsfonts}
\usepackage{algorithmic}
\usepackage{graphicx}
\usepackage{textcomp}
\usepackage{multirow}
\usepackage[font=small]{caption}
\usepackage{subcaption}
\usepackage[table,xcdraw]{xcolor}
\usepackage{hyperref}
\usepackage{booktabs}
\usepackage{balance}
\usepackage{tcolorbox} 
\usepackage{xcolor} 
\IEEEoverridecommandlockouts                              

\usepackage{graphicx}

\title{\LARGE \bf
PhysicalAgent: Towards General Cognitive Robotics with Foundation World Models
}

\author{Artem Lykov$^{*}$, Jeffrin Sam$^{*}$, Hung Khang Nguyen$^{*}$, Vladislav Kozlovskiy$^{*}$, Yara Mahmoud,\\
Valerii Serpiva, Miguel Altamirano Cabrera, Mikhail Konenkov, Dzmitry Tsetserukou%
\thanks{$^{*}$Denotes equal contribution.}%
\thanks{All authors are with the Intelligent Robotics Laboratory, Skolkovo Institute of Science and Technology (Skoltech), 
Bolshoy Boulevard 30, bld. 1, Moscow 121205, Russia. 
{\tt\small \{Artem.Lykov, Jeffrin.Sam, Khang.Nguyen, Vladislav.Kozlovskiy, Yara.Mahmoud, Valerii.Serpiva, M.Altamirano, Mikhail.Konenkov, D.Tsetserukou\}@skoltech.ru}}%
}

\begin{document}

\maketitle
\thispagestyle{empty}
\pagestyle{empty}

\begin{abstract}

We introduce \textbf{PhysicalAgent}, an agentic framework for robotic manipulation that integrates iterative reasoning, diffusion-based video generation, and closed-loop execution. Given a textual instruction, our method generates short video demonstrations of candidate trajectories, executes them on the robot, and iteratively re-plans in response to failures. This approach enables robust recovery from execution errors. We evaluate PhysicalAgent across multiple perceptual modalities (egocentric, third-person, and simulated) and robotic embodiments (bimanual UR3, Unitree G1 humanoid, simulated GR1), comparing against state-of-the-art task-specific baselines. Experiments demonstrate that our method consistently outperforms prior approaches, achieving up to 83\% success on human-familiar tasks. Physical trials reveal that first-attempt success is limited (20–30\%), yet iterative correction increases overall success to 80\% across platforms. These results highlight the potential of video-based generative reasoning for general-purpose robotic manipulation and underscore the importance of iterative execution for recovering from initial failures. Our framework paves the way for scalable, adaptable, and robust robot control.

\end{abstract}


\section{Introduction}

The rapid progress of large foundation models has transformed the design of AI agents. Unlike prompt-only systems, modern agents \cite{manus_shen2025,owl_hu2025,operator_openai2025} demonstrate autonomy by reasoning ahead, decomposing complex problems, and iteratively acting through tools. A central factor behind this shift is the reliance on general-purpose \emph{foundation models}, which reduce dependence on task-specific training, enable seamless model updates as new versions are released, and exploit economies of scale from shared infrastructure. This paradigm has led to flexible and high-performing agents in the virtual domain.

\begin{figure}[thpb]
    \centering
    \includegraphics[width=0.47\textwidth]{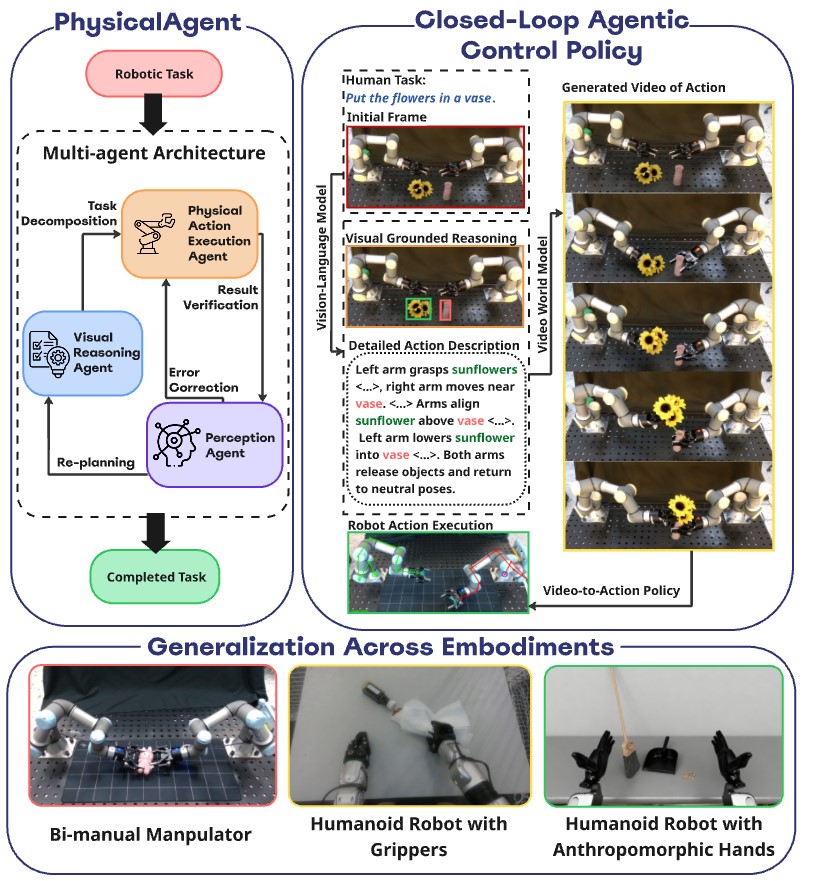}
    \caption{\textbf{PhysicalAgent:} Our framework enables general cognitive robotics by combining vision--language reasoning with a diffusion-based foundation world model that generates future action videos. Unlike prior VLA systems that require retraining for new robots or environments, PhysicalAgent remains model-agnostic: only a lightweight robot-specific adapter maps generated videos into motor commands using onboard feedback. This design allows seamless transfer across heterogeneous embodiments, demonstrated on bi-manual manipulation and humanoid control in simulation and physical robot.}
    \label{fig:teaser}
    \vspace{-2mm}
\end{figure}

Extending such agentic capabilities into the physical world, however, poses unique challenges. Robotic systems must ground reasoning in perception, operate across heterogeneous embodiments, and execute reliably in dynamic environments. Existing approaches in cognitive robotics, such as RT-1/RT-2 \cite{brohan2022rt1, brohan2023rt2}, OpenVLA \cite{openvla_kim2024}, and Isaac-Gr00t \cite{bjorck2025gr00t}, rely on vision–language–action (VLA) models that require task- and platform-specific fine-tuning. While effective on their training distributions, these methods remain brittle when transferred to new robots, environments, or tasks. Similarly, recent work on world-model-based action synthesis \cite{dreamgen_jang2025}, has demonstrated compelling closed-loop execution but depends on specialized models (e.g., Cosmos Predict \cite{agarwal2025cosmos}) trained on carefully curated robot–scene pairs, limiting generality.

In this work, we introduce \emph{PhysicalAgent}, a general framework for cognitive robotics with foundation world models. PhysicalAgent is designed around a \emph{Perceive \textrightarrow{} Plan \textrightarrow{} Reason \textrightarrow{} Act} pipeline, where first two leverages foundation models that require no robot-specific training, and only the actor part should be trained. Perception and grounded reasoning are enabled by vision–language models that translate natural language goals into structured, constraint-aware execution plans \cite{vlm_reasoning_zhang2024}. Action generation is realized through a diffusion-based foundation world model that synthesizes future trajectories or action videos. To execute these synthesized behaviors on real robots, we train a lightweight embodiment-specific policy that maps generated video actions into low-level motor commands using onboard camera feedback. This ensures that the same high-level pipeline can generalize across heterogeneous robotic platforms while only requiring minimal per-robot adaptation.

We evaluate PhysicalAgent on both simulated and real-world tasks, spanning mobile navigation and long-horizon manipulation, and demonstrate three key properties: (i) robustness to environmental variability, (ii) transferability across embodiments, and (iii) scalability through foundation-model-driven reasoning and acting. By releasing the system, interfaces, and evaluation protocols as open source, we aim to establish PhysicalAgent as a platform for reproducible and extensible research in cognitive robotics.


\section{Related Work} 
\label{sec:related_work} 

Our work builds upon recent advances in agentic AI, foundation models, and generative action synthesis for robotics. We situate \emph{PhysicalAgent} relative to prior research in these domains. 

\subsection{Cognitive Architectures and Agentic AI in Robotics} 

The pursuit of autonomous robots has long been guided by cognitive architectures that structure perception, planning, and execution \cite{laird2019soar}. Traditional pipelines often relied on symbolic planners and manually engineered state representations, which limited their adaptability to unstructured environments. The advent of large language models (LLMs) has catalyzed a paradigm shift, enabling agents to reason over complex, open-ended goals expressed in natural language and even orchestrate entire manufacturing pipelines with swarms of heterogeneous robots \cite{lykov2024industry}.

Early integrations of LLMs in robotics, such as SayCan \cite{ahn2022saycan}, demonstrated how models can ground high-level instructions into a sequence of executable skills by scoring them against affordances perceived in the environment. Subsequent works like RT-1 and RT-2 \cite{brohan2022rt1, brohan2023rt2} moved towards end-to-end visuomotor control, tokenizing robot actions and co-fine-tuning language models on vast datasets of web and robotics data to create generalist policies. Similarly, PaLM-E \cite{driess2023palm} showed that multimodal language models can directly ingest sensor data to produce textual plans or low-level commands. While powerful, these approaches often result in monolithic policies that entangle high-level reasoning with low-level control, making them data-intensive to train and difficult to adapt across embodiments. Recognizing these limitations, other works have explored modular frameworks; for example, CognitiveOS presents a multi-agent operating system designed to endow different types of robots with cognitive capabilities through configurable modules \cite{lykov2024cognitiveos}.

In contrast, \emph{PhysicalAgent} adopts a modular, decoupled architecture. Our \emph{Perceive \textrightarrow{} Plan \textrightarrow{} Reason \textrightarrow{} Act} pipeline explicitly separates high-level reasoning from low-level execution. This modularity enhances portability and allows for explicit monitoring and re-planning, a feature critical for robust long-horizon task execution. Recent work has also focused on improving long-horizon performance within end-to-end VLA models; for instance, Long-VLA introduces a phase-aware input masking strategy to enhance subtask compatibility and skill chaining \cite{fan2025longvla}. Other approaches, such as CogVLA, focus on improving the computational efficiency and scalability of VLA models through cognition-aligned strategies like instruction-driven routing and token sparsification \cite{li2025cogvla}.

\subsection{Foundation Models for Robotic Perception and Reasoning} 

Vision-Language Models (VLMs) have become a cornerstone of modern robotics, offering unprecedented capabilities for zero-shot scene understanding and visual grounding \cite{radford2021clip, alayrac2022flamingo}. In robotics, VLMs are widely used for task decomposition, object identification, and affordance detection. For instance, some systems use VLMs to parse an initial instruction into a sequence of subgoals \cite{huang2022language} or to serve as a semantic planner for complex bimanual manipulation tasks \cite{gbagbe2024bivla}. Recent work seeks to fuse the semantic understanding of VLMs with the predictive power of world models to achieve more robust, physics-aware task planning \cite{wang2023robogen}.

Our work leverages VLMs not as a one-shot planner but as a continuous cognitive engine. As detailed in Section~\ref{sec:visual_reasoning}, \emph{PhysicalAgent} invokes visual reasoning at multiple stages: initial task decomposition, generating constraint-aware descriptions for action synthesis, and monitoring execution outcomes. This iterative, closed-loop use of VLMs for reasoning and correction distinguishes our approach from systems that primarily use them for open-loop planning. While we employ models like Gemini Pro \cite{team2023gemini}, our framework is model-agnostic, designed to readily incorporate more powerful VLMs as they become available \cite{openai2023gpt4, achiam2023gpt, anil2024claude, bai2024qwen}.

\subsection{Action Synthesis from Generative World Models} 

The dominant paradigms for robotic control have been reinforcement learning (RL) and imitation learning (IL). While RL can discover novel policies, it is often sample-inefficient and struggles with complex, long-horizon tasks. IL, including methods like behavior cloning, is more direct but requires large datasets of expert demonstrations specific to a single robot embodiment, limiting scalability.

Recent work has explored generative models to overcome these limitations. Diffusion policies, for example, learn to generate action trajectories by reversing a diffusion process, showing promising results in complex manipulation tasks \cite{chi2023diffusion, reuss2023goal}. This paradigm is rapidly evolving, with recent works exploring discrete diffusion for action decoding to better align with the token-based interfaces of VLMs \cite{liang2025discrete}. Other approaches have focused on learning from diverse, passive data sources like human videos, with multimodal transformers showing strong performance in general manipulation tasks \cite{mees2024vima}, but still face challenges in bridging the morphology gap between humans and robots.

\emph{PhysicalAgent} introduces a novel intermediate representation for action synthesis: generated video. We treat a text-to-video diffusion model as a general-purpose, embodiment-agnostic world model \cite{ho2022imagen, singer2022make}. Instead of directly outputting robot torques or end-effector poses, our system generates a physically plausible video of the desired subtask. This approach leverages the rich, implicit understanding of physics and object interactions learned by state-of-the-art video models from massive web-scale data. The final step—mapping this video to motor commands via a lightweight pose-to-action adapter—becomes a much simpler, supervised learning problem that requires minimal data collection per embodiment. This philosophy aligns with emerging research on adapting large pre-trained models by leveraging vast, diverse robotics datasets collected across multiple embodiments \cite{rtx_2023}. This unique factorization of the problem allows \emph{PhysicalAgent} to inherit the rapid progress of the video generation field while maintaining a clear and efficient path to deployment on diverse physical hardware.


\section{PhysicalAgent Architecture} 

As a core architectural principle, our agent leverages perception and reasoning modules that are not dependent on a specific embodiment. The only embodiment-specific component in our pipeline is a lightweight skeleton detection model, which is neither computationally intensive nor data-demanding. In this section, we provide a detailed rationale for this architectural choice and demonstrate its embodiment-agnostic nature.

The field of video generation \cite{li2023videogen} is developing at a rapid pace and has become a highly active research area. State-of-the-art models are pretrained on vast amounts of multimodal data, including large-scale video corpora and first-person recordings of diverse activities. As a result, these models acquire an implicit understanding of fundamental physical processes and everyday interactions with the world. Moreover, video generation models are increasingly accessible through API-based interfaces, which enables rapid prototyping of pipelines, seamless integration of newly released models, and continuous improvement in output quality without the need for local training or hosting. This combination of scalability, accessibility, and generalization makes video generation models particularly attractive for robotics applications.

In addition, video generation models are trained to align text prompts with corresponding visual outputs. This mechanism parallels the way humans reason about actions: by conceptualizing an instruction and then mentally simulating its execution. Crucially, these models can extend such reasoning to robotic agents, imagining how specific actions should be performed, without requiring any prior knowledge of the robot’s internal architecture. Text-based instructions, which fall squarely within the training distribution of these models, thus provide a natural and effective interface for specifying robotic behavior.

Finally, to adapt the generated video outputs to robot control, we only require a lightweight model that maps visual representations to robot states. Constructing such a dataset and training this mapping is significantly simpler than training reinforcement learning policies from scratch. The training process is more stable, requires far less data, and can be efficiently supported by existing computer vision foundation models \cite{wang2024yolov10, yolo11} (e.g., YOLO-based architectures).

\begin{figure}[t]
    \centering
    \includegraphics[width=0.95\linewidth]{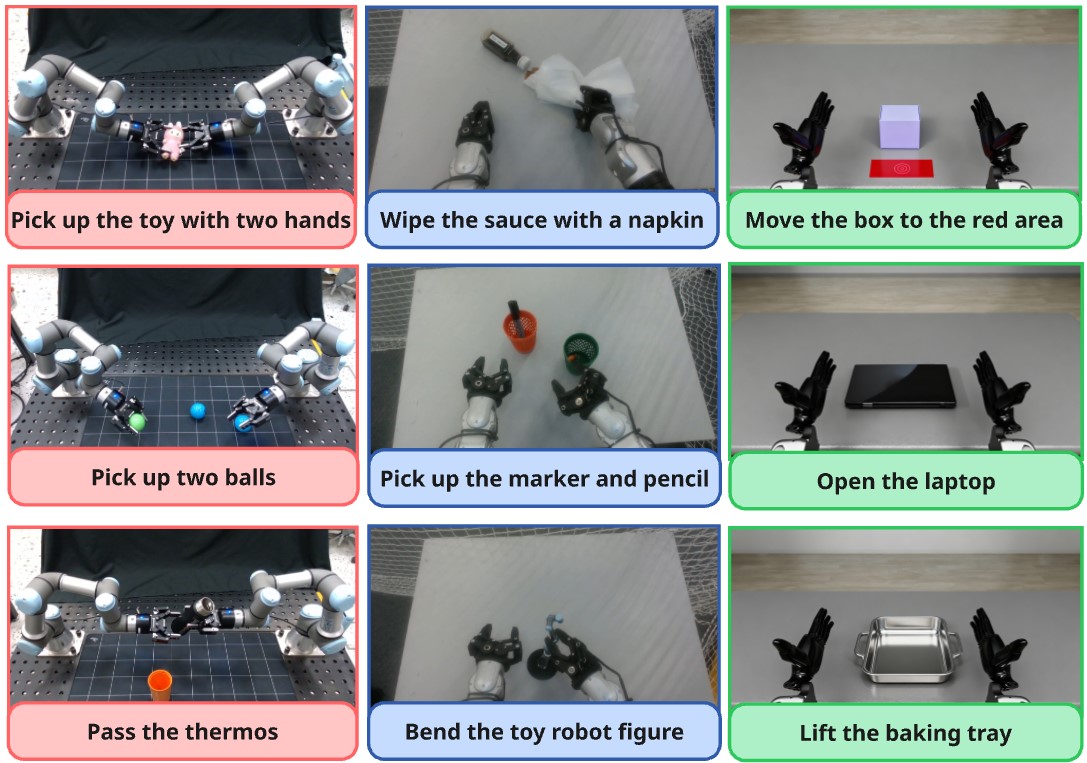}
    \caption{
    \textbf{Zero-shot video generation across multiple tasks and embodiments.} 
    Our diffusion-based world model generates task videos without any retraining or fine-tuning for specific robots. 
    The figure illustrates diverse manipulation tasks on three embodiments: a bimanual UR3 setup (left), a Unitree G1 humanoid (center), and a GR1 humanoid robot in simulation (right). 
    This demonstrates that the high-level reasoning and video generation pipeline is embodiment-agnostic and generalizes well to unseen tasks and robots.
    }
    \vspace{-2mm}
    \label{fig:task_gallery}
\end{figure}

Importantly, this architecture enables cross-embodiment generalization: the same perception–reasoning pipeline can generate feasible task rollouts for entirely different robotic morphologies without retraining or fine-tuning. Figure~\ref{fig:task_gallery} illustrates this capability by showing example rollouts for diverse manipulation tasks executed on three distinct embodiments: a bimanual UR3 setup, a Unitree G1 humanoid, and a simulated dual-arm agent. This demonstrates that our approach provides a unified substrate for reasoning and action planning that remains valid across heterogeneous robotic platforms.



\begin{figure*}[!t]
    \vspace{-2mm}
    \centering
    \includegraphics[width=0.84\linewidth]{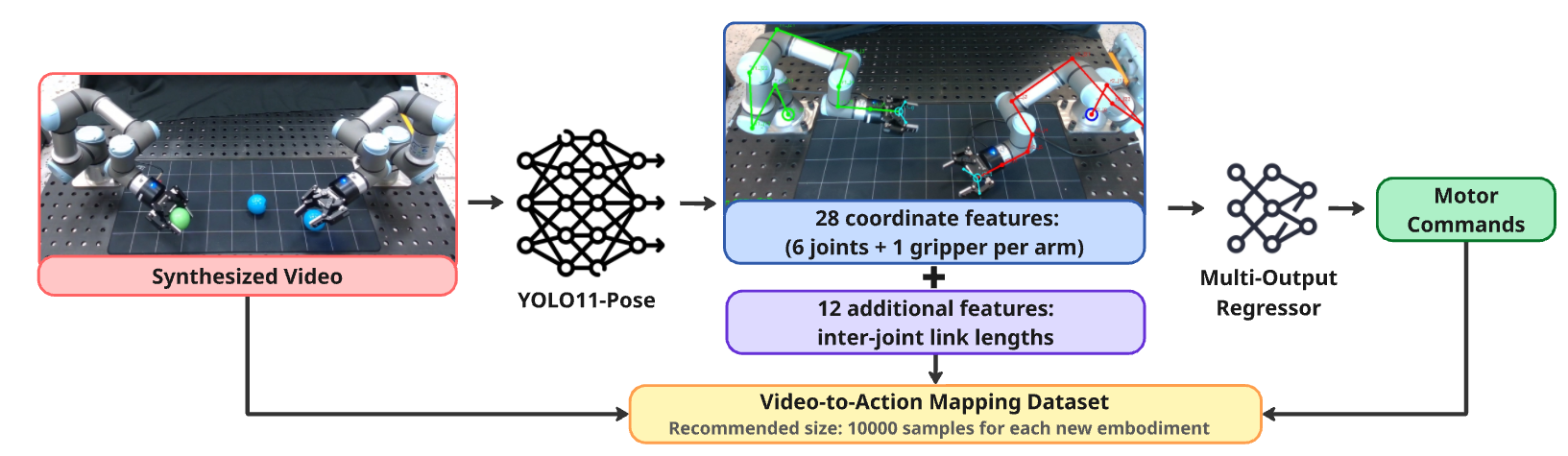}
    \caption{Embodiment-specific skill execution pipeline. Left-to-right flow: synthesized video → pose extraction → feature computation → regression → robot execution. Example shows a bimanual manipulator with a third-person camera.}
    \label{fig:skill_pipeline}
    \vspace{-2mm}
\end{figure*}

\section{Grounded Visual Reasoning With Vision--Language Models}
\label{sec:visual_reasoning}

A crucial step in executing a robotic task is understanding the goal and grounding it in perception. In PhysicalAgent, this is achieved through repeated use of visual reasoning powered by Vision Language Models (VLMs). Unlike task-specific perception modules, foundation VLMs provide general-purpose scene understanding, making them particularly suited for heterogeneous robots and diverse tasks.

\paragraph{Task Understanding and Decomposition}
Immediately after receiving a natural-language goal and perceiving the initial scene, the agent performs visual reasoning to transform the high-level instruction into a structured plan. This process involves decomposing the task into a sequence of atomic subtasks or skills (e.g., ``grasp object,'' ``stabilize container,'' ``insert into slot''). These atomic skills serve as the interface between abstract planning and concrete execution.

\paragraph{Contextualized Scene Descriptions}
For each subtask, the agent must generate a detailed and constraint-aware prompt that will be passed to the action generation module. This requires not only identifying relevant objects and their spatial relations but also capturing physical constraints and robot-specific affordances. The VLM is used to ground this context in a textual description that conditions the diffusion-based world model to synthesize feasible action trajectories.

\paragraph{Execution Monitoring and Correction}
Visual reasoning is invoked again after each execution step to assess progress and correct errors. Given the original task and a pair of images (before and after execution), the VLM evaluates whether the subtask was successfully completed and recommends one of three outcomes: (i) continue with the plan, (ii) retry the current action, or (iii) return to high-level planning. This closed-loop use of reasoning ensures robustness in long-horizon tasks.

\paragraph{Role of Foundation Models}
Visual reasoning thus acts as the cognitive backbone of the PhysicalAgent pipeline, enabling robots to interpret goals, ground them in the environment, and adapt dynamically to execution outcomes. Importantly, our design is model-agnostic: any modern VLM capable of multi-image reasoning can be used. In our implementation we primarily employed \emph{Gemini Pro Flash}~\cite{team2023gemini}, but other models such as GPT-4o~\cite{openai2023gpt4}, Claude-3.5 Sonnet~\cite{anil2024claude}, or QwenVL~\cite{bai2024qwen} are equally applicable.

\section{Diffusion-based World Models for Action Generation}

At the heart of our framework lies a diffusion-based foundation world model, which reconceptualizes action generation as conditional video synthesis rather than direct policy learning. This design leverages large-scale image-to-video foundation models that, when conditioned on a single first frame and a short textual description, produce physically plausible short rollouts capturing object dynamics, contact events, and scene-level causal structure. Importantly, prior work such as DreamGen \cite{dreamgen_jang2025} demonstrates the promise of world-model-driven action synthesis but depends on specialized prediction models (e.g., Cosmos Predict \cite{agarwal2025cosmos}) trained on carefully curated robot–scene pairs, which limits out-of-distribution generalization and increases the engineering burden for deployment. In contrast, we show that off-the-shelf foundation image→video models — originally trained for general image or text conditioned video synthesis — can serve as practical and effective diffusion world models for embodied agents.

Concretely, our pipeline uses a foundation image-to-video generator to synthesize candidate rollouts from the robot's current camera observation and an instruction or plan. These synthesized videos act as an intermediate, human- and machine-interpretable modality for downstream modules: (i) a lightweight video-to-control adapter that maps visualized trajectories to executable motor commands using onboard sensors and closed-loop feedback; and (ii) verifier modules that evaluate feasibility and safety before execution. By offloading dynamics priors to the generative model, the robot-specific component is reduced to a small adapter that requires far less task- or platform-specific data.

This design brings three practical advantages. First, it significantly lowers the data and engineering burden when deploying to new robots and environments because the costly prior (the generative model) is task- and embodiment-agnostic. Second, it permits rapid upgrades: when a superior image-to-video foundation becomes available, it can be swapped in without retraining the adapter. Third, synthesized rollouts provide an inspectable layer for monitoring and human-in-the-loop correction, improving transparency and safety.

We emphasize that modern image-to-video systems already support first-frame + description conditioning — examples include Seedance 1.0 Pro \cite{seedance2024}, Google Veo 2\cite{veo2_google2024}, Luma Ray2\cite{luma_ray2_2025}, and Wan 2.2\cite{wan22_replicate2025} — and can be accessed via commercial APIs or lightweight open-source variants. In our experiments we use the Wan 2.2 Image-to-Video Fast model for its favorable speed/quality trade-off, but the architecture and training of our adapter make the approach model-agnostic: any compatible image-to-video foundation can be substituted as improvements arrive.

By grounding robotic control in diffusion-driven video generation, \emph{PhysicalAgent} establishes a scalable, embodiment-agnostic mechanism for action synthesis and validation — closing the gap between foundation-model reasoning and robust physical execution.

\begin{table*}[!t]
\caption{Comparison with existing approaches for bimanual UR3. Our method shows highest success rates.}
\centering
\resizebox{\textwidth}{!}{%
\begin{tabular}{|l|c|c|c|c|c|c|c|c|c|c|c|c|c|}
\hline
Method & (a) box & (b) ball & (c) buttons & (d) plate & (e) drawer & (f) fridge & (g) handover & (h) laptop & (i) rope & (j) dust & (k) tray & (l) handover easy & (m) oven \\
\hline
ACT                   & 0\%  & 36\% & 4\%  & 0\%  & 13\% & 0\%  & 0\%  & 0\%  & 16\% & 0\%  & 6\%  & 0\%  & 2\%  \\ \hline
RVT-LF                & 52\% & 17\% & 39\% & 3\%  & 10\% & 0\%  & 0\%  & 3\%  & 3\%  & 0\%  & 6\%  & 0\%  & 3\%  \\ \hline
PerAct-LF             & 57\% & 40\% & 10\% & 2\%  & 27\% & 0\%  & 0\%  & 11\% & 21\% & 28\% & 14\% & 9\%  & 8\%  \\ \hline
PerAct2               & 6\%  & 50\% & 47\% & 4\%  & 10\% & 3\%  & 11\% & 12\% & 24\% & 0\%  & 1\%  & 41\% & 9\%  \\ \hline
PhysicalAgent (ours) & 42\% & 55\% & 73\% & 12\% & 24\% & 31\% & 40\% & 6\%  & 37\% & 10\% & 6\%  & 62\% & 43\% \\ \hline
\end{tabular}%
}

\label{tab:results}
\end{table*}

\begin{table*}[!t]
\caption{Performance of different robotic platforms across tasks. G1 humanoid shows highest median success rates.}
\centering
\resizebox{\textwidth}{!}{%
\begin{tabular}{|l|c|c|c|c|c|c|c|c|c|c|c|c|c|}
\hline
Platform & (a) box & (b) ball & (c) buttons & (d) plate & (e) drawer & (f) fridge & (g) handover & (h) laptop & (i) rope & (j) dust & (k) tray & (l) handover easy & (m) oven \\
\hline
Bimanual              & 42\% & 55\% & 73\% & 12\% & 24\% & 31\% & 40\% & 6\%  & 37\% & 10\% & 6\%  & 62\% & 43\% \\ \hline
G1 humanoid           & 51\% & 57\% & 83\% & 20\% & 42\% & 51\% & 46\% & 18\% & 45\% & 31\% & 18\% & 65\% & 62\% \\ \hline
GR1 humanoid (sim)    & 37\% & 22\% & 67\% & 14\% & 20\% & 31\% & 27\%  & 13\% & 31\% & 14\% & 6\% & 62\% & 47\%  \\ \hline
\end{tabular}%
}
\label{tab:platform-results}
\end{table*}

\section{Embodiment-specific Skill Execution From Generated Videos}

The overall pipeline for embodiment-specific skill execution is illustrated in Fig.~\ref{fig:skill_pipeline}. The diagram shows the flow from synthesized subtask videos through YOLO11-Pose keypoint extraction, feature computation, and regression to motor commands, culminating in real robot execution. The example demonstrates a bimanual manipulator with a third-person camera view.

In the previous stages of the PhysicalAgent pipeline, we generated short videos of subtask execution from a single initial frame and a natural-language description, ensuring that the resulting trajectories were executable and visually correct. The final stage—mapping these synthesized videos to real robot actions—is the only part of the pipeline that requires platform-specific adaptation.

We decompose this task into two main components. First, we estimate the 2D positions of key robot joints in each frame of the synthesized video using a lightweight pose detection model (a fine-tuned YOLO11-Pose \cite{yolo11}). For bimanual manipulators, we tracked 14 keypoints (6 joints + 1 gripper per arm), producing 28 coordinate features. To enhance the representation, we computed inter-joint link lengths (12 features), resulting in a final 40-dimensional feature vector per frame. Missing keypoints due to occlusions or detection failures were imputed using a simple mean strategy (SimpleImputer).

Second, these features are fed into a regression model to predict the corresponding low-level motor commands. We employ a \texttt{MultiOutputRegressor} wrapping a \texttt{HistGradientBoostingRegressor} with the following hyperparameters: \texttt{max\_iter = 500}, \texttt{learning\_rate=0.1}, \texttt{min\_samples\_leaf=20}, \texttt{early\_stopping = True}. This procedure yields accurate motor commands with mean absolute errors below 0.1 for most joints; larger errors appear in joints with wide motion ranges (e.g., J7).

The training dataset for this adapter consisted of approximately 10,000 samples, collected in about 30 minutes at a rate of 5 samples per minute. Each sample contains the camera frame corresponding to the perception input, the robot's current joint positions, and the projected 2D keypoints derived from the camera's intrinsics and inverse kinematics. This lightweight approach eliminates the need for fine-tuning large models while supporting onboard inference on consumer-grade hardware. The pipeline is compatible with both third-person and egocentric camera views.

Pose estimation was performed using a fine-tuned YOLO11-Pose model \cite{yolo11}, pretrained and trained for 150 epochs with batch size 8, image size 1280, and early stopping with patience 50. Optimizer parameters included \texttt{lr0=0.01}, \texttt{momentum=0.937}, \texttt{weight\_decay=0.0005}, and warm-up for 3 epochs. Training utilized 4 workers on an RTX 4090, and the first 10 layers of the backbone were frozen to preserve pretrained representations (\texttt{freeze=10}). Data augmentation was minimal to preserve geometric consistency, and masks were applied for occluded joints. Validation was conducted on held-out frames to ensure accurate keypoint localization across different camera views.

This pipeline allows for straightforward adaptation to new robot platforms. Only the pose regression component needs retraining with a small dataset collected on the target robot. The architecture of the adapter remains identical, making it lightweight, fast to train, and compatible with different kinematic structures. The approach has been successfully applied to both humanoid and bimanual manipulators, and all scripts, sample datasets, and trained models are available in the open-source repository.

\section{Experimental Evaluation}

The goal of our experimental study is twofold:  
(1) to validate the generality of the proposed architecture across different perceptual modalities and robot embodiments, and  
(2) to quantify the benefit of the iterative agentic control loop in improving task success rate under real-world execution.

To this end, we design two complementary experiments.  
The first experiment (Section~\ref{sec:exp1}) explores how different observation modalities affect task execution (simulation, egocentric view, third-person view).  
The second experiment (Section~\ref{sec:exp2}) focuses purely on physical robot trials on two distinct platforms, measuring success rate and number of corrective iterations.

\subsection{Experiment 1: Embodiment and Perception Study}
\label{sec:exp1}

\subsubsection{Objective}  
Evaluate whether PhysicalAgent achieves higher task success rates than task-specific baselines, and assess generalization across different embodiments.

\subsubsection{Hypotheses}  
\begin{itemize}
  \item[\textit{(H1)}] Our agent achieves a higher mean success rate compared to four baseline methods: ACT, RVT-LF, PerAct-LF, and PerAct2.
  \item[\textit{(H2)}] Performance is highest on the G1 humanoid due to alignment with human-centric data; we expected lower performance on the bimanual UR3 setup and the simulated GR1 humanoid due to kinematic and domain mismatches with the training data.
\end{itemize}

\subsubsection{Experimental Setup}  
Three platforms were used: a bimanual UR3 setup, a Unitree G1 humanoid, and a simulated GR1 humanoid. Thirteen manipulation tasks were evaluated: box, ball, buttons, plate, drawer, fridge, handover, laptop, rope, dust, tray, handover easy, and oven.  
For each task-platform pair, 30 videos were generated and executed; success was scored as binary.

\subsubsection{Metrics}  
Task-level success rate (mean $\pm$ 95\% CI).  

\subsubsection{Statistical Analysis}  
ANOVA was performed to test the effect of the implemented \textit{method} on the success rate for H1 and the effect of the used \textit{platform} for H2.

\begin{figure}[!t]
    \centering
    \vspace{-2mm}
    \includegraphics[width=0.93\linewidth]{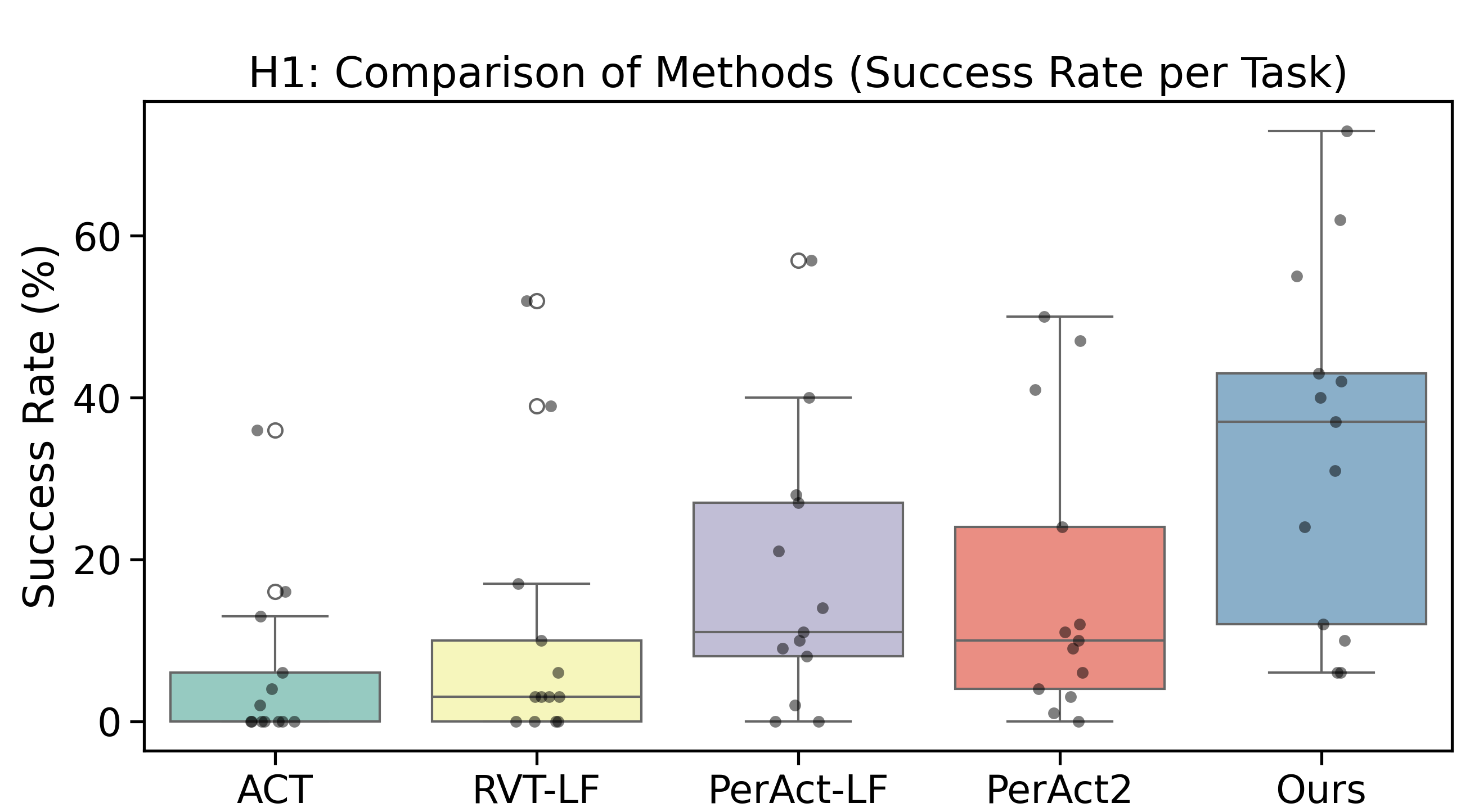}
    \caption{Mean task success rate across different methods (H1). Our diffusion-based agent significantly outperforms task-specific baselines, as confirmed by ANOVA ($F(4,60)=5.04, p=0.0014$). Error bars indicate 95\% confidence intervals.}
    \label{fig:h1-results}
    \vspace{-4mm}
\end{figure}

\subsubsection{Results}  
\begin{figure}[!t]
    \centering
    \includegraphics[width=0.79\linewidth]{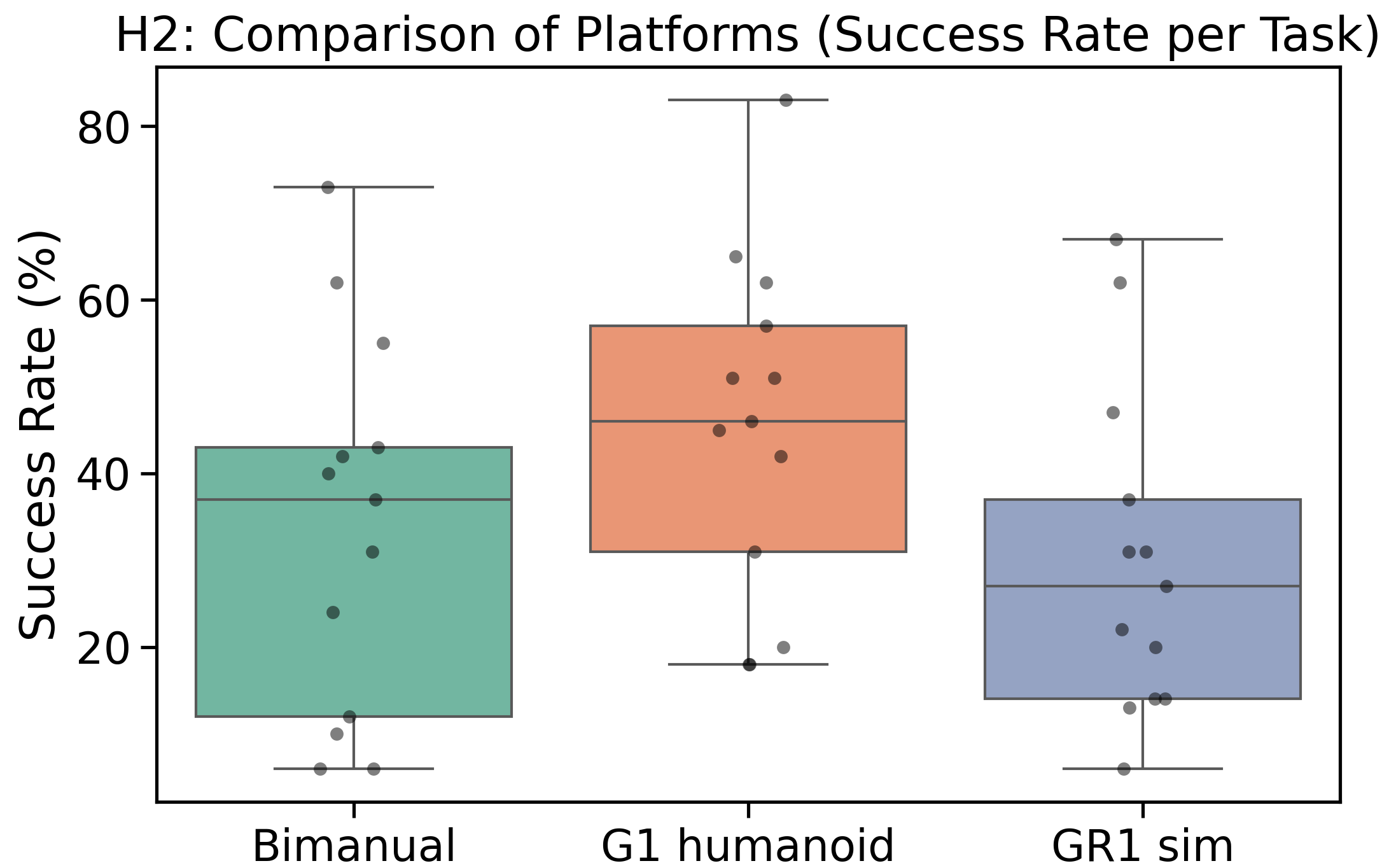}
    \caption{Task success rate distribution across robotic platforms (H2). The G1 humanoid achieves the highest median performance, though differences are not statistically significant ($F(2,36)=2.01, p=0.1485$). Boxplots indicate interquartile ranges and median values.}
    \label{fig:h2-results}
\end{figure}

\begin{itemize}
\item \textbf{Hypothesis H1: Method Effect.}  
ANOVA shows a statistically significant effect of the chosen \textit{Method} on the task success rate ($F(4,60)=5.04, p=0.0014$), indicating that our diffusion-based agent significantly outperforms the baselines (see Fig.~\ref{fig:h1-results}), confirming H1.

\item \textbf{Hypothesis H2: Platform Effect.}  
ANOVA indicates no statistically significant effect of the \textit{Platform} on perormance ($F(2,36)=2.01, p=0.1485$). Although G1 shows the highest median success rates (Fig.~\ref{fig:h2-results}), differences are not statistically significant at $\alpha=0.05$. This suggests that while embodiment alignment helps, variability across tasks reduces significance, leading us to reject our second hypothesis (H2).

\end{itemize}

\subsection{Experiment 2: Iterative Physical Task Execution}
\label{sec:exp2}

\subsubsection{Objective}  

This experiment was designed to evaluate whether our iterative \emph{Perceive \textrightarrow{} Plan \textrightarrow{} Reason \textrightarrow{} Act} pipeline could achieve high task success rates in a physical setting and to determine if this performance would generalize across different robot embodiments.

\subsubsection{Hypotheses}  
\begin{itemize}
  \item[(H1)] The proposed agentic architecture will achieve a high final success rate (significantly greater than that of a single-shot execution), regardless of of the robot's embodiment (UR3 bimanual setup vs Unitree G1 humanoid).  
  \item[(H2)] Most successful task completions would require at least one corrective iteration; specifically, fewer than 50\% of successes will occur on the first attempt.
\end{itemize}

\subsubsection{Experimental Setup}  
\begin{itemize}
  \item \textbf{Platforms:} two physical robots — (i) a bimanual UR3 setup, and (ii) a Unitree G1 humanoid.  
  \item \textbf{Tasks:} a fixed set of 10 manipulation tasks (same tasks on both platforms).  
  \item \textbf{Procedure:} for each task, the agent followed a standardized procedure: (1) generates a plan + video from text instruction; (2) executes on robot; (3) evaluates outcome; (4) if failure and recoverable, re‐plans. Termination if success, an irrecoverable failure (e.g., dropping an object out of reach), or after 10 attempts.  
  \item \textbf{Repetitions:} each of the 10 tasks was run once per platform, resulting in 20 total experimental runs.
\end{itemize}

\subsubsection{Metrics}  
\begin{itemize}
  \item Final success rate per platform (the proportion of tasks successfully completed out of 10).  
  \item First‐attempt success rate (how many tasks succeeded on the first try).  
  \item Number of iterations required for each successful task.  
  \item Survival curve: fraction of tasks still unsolved after $n$ iterations.  
\end{itemize}

\begin{table}[t]
\centering
\caption{Per-task results for both physical platforms. A dash indicates that the task did not succeed within the maximum allowed attempts. First-attempt successes are highlighted in bold.}
\label{tab:task_results}
\begin{tabular}{l|c|c|c|c}
\hline
\multirow{2}{*}{\textbf{Task}} & \multicolumn{2}{c|}{\textbf{Bimanual (UR3)}} & \multicolumn{2}{c}{\textbf{Humanoid (G1)}} \\ 
 & Res. & Iter. & Res. & Iter. \\ \hline
Fold the tissue         & F & -- & F & -- \\
Wipe the liquid         & S & 3  & S & 4  \\
Pick the ball           & S & 4  & S & 5  \\
Lift the baking tray    & \textbf{S} & \textbf{1}  & S & 3  \\
Pick the thermos        & S & 3  & S & 2  \\
Put a pencil in a glass & S & 2  & S & 3  \\
Pick two balls          & S & 3  & S & 3  \\
Pick the toy            & F & -- & F & -- \\
Push the cube           & \textbf{S} & \textbf{1} & \textbf{S} & \textbf{1} \\
Pick the cube           & \textbf{S} & \textbf{1} & \textbf{S} & \textbf{1} \\ \hline
\textbf{Final Success Rate}        & \multicolumn{2}{c|}{\textbf{8/10 (80\%)}} & \multicolumn{2}{c}{\textbf{8/10 (80\%)}} \\
\textbf{First-Attempt Success Rate} & \multicolumn{2}{c|}{\textbf{3/10 (30\%)}} & \multicolumn{2}{c}{\textbf{2/10 (20\%)}} \\
\textbf{Mean Iterations (Success)} & \multicolumn{2}{c|}{\textbf{2.25}} & \multicolumn{2}{c}{\textbf{2.75}} \\ \hline
\end{tabular}
\vspace{-2mm}
\end{table}

\subsubsection{Results}  
Table~\ref{tab:task_results} shows per‐task outcomes and iteration counts to success. First‐attempt successes are highlighted in bold. Both platforms achieved a final success rate of 80\%, while only 30\% of UR3 tasks and 20\% of G1 tasks succeeded on the first attempt. This demonstrates the clear benefit of the iterative reasoning–execution loop, which allows recovery from initial failures and substantially increases overall task completion rate.

\begin{figure}[!t]
    \vspace{-2mm}
    \centering
    \includegraphics[width=0.9\linewidth]{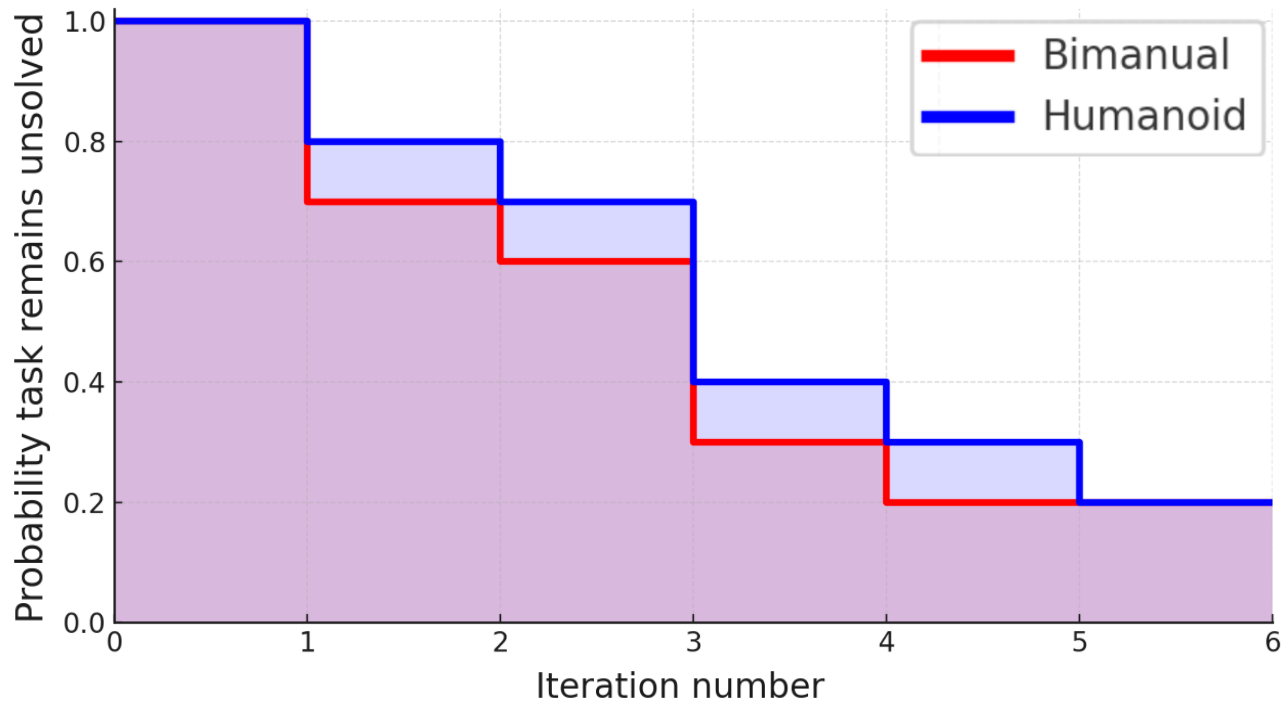}
    \caption{Kaplan–Meier–style survival curves showing the fraction of tasks that remain unsolved after $n$ iterations for the two physical platforms. Bold solid lines show the survival estimate (red - Bimanual (UR3); blue — Humanoid (Unitree G1)).}
    \vspace{-3mm}
    \label{fig:survival_curve}
\end{figure}

The Kaplan-Meier-style survival curves in Fig.~\ref{fig:survival_curve} illustrate the fraction of unsolved tasks remaining after each corrective iteration. For both platforms, the curves show a sharp decline within the first three iterations, with convergence by iteration 4 for nearly all solvable tasks.

\subsubsection{Discussion}  
These findings support both hypotheses: the agentic, iterative loop delivered a high overall success rate of 80\% on both platforms (H1), a substantial improvement over the low first-attempt success rates of 30\% and 20\% (H2), highlighting the benefit of multiple corrective iterations. Performance generalizes across embodiments, demonstrating robustness of the iterative reasoning–execution approach. Remaining failures are associated with irrecoverable errors in execution, such as object drops or hardware constraints, pointing to future work on enhanced recovery strategies.

\section{Limitations and Social Impact} 
\textbf{Limitations.}
While the proposed PhysicalAgent system demonstrates significant potential, it is important to acknowledge its current limitations, which provide clear directions for future research.

The tasks performed successfully, while involving complex trajectory generation, represent relatively simple manipulation challenges.The system struggled with tasks involving the manipulation of deformable objects. This highlights a current boundary in the system's ability to reason about and interact with objects that lack rigid-body physics. Future work could address this by incorporating more sophisticated world models that can simulate soft-body dynamics or by integrating tactile feedback to allow the agent to react to material properties in real time.

Furthermore, the system's reliance on iterative correction, while a core strength for overall robustness, points to a limitation in first-attempt reliability. The initial success rates of 20-30\% indicate that the system is better framed as robust through repetition rather than as a consistently ``first-time-right" solution. This limits its immediate applicability in time-critical scenarios. Future research should focus on enhancing the quality of the initial video generation and developing more advanced re-planning heuristics to increase the probability of success on the first attempt.

Finally, the computational cost associated with generating trajectories is substantial. A generation time of up to 30 seconds for a five-second action video currently constrains real-time applicability and scalability. For the system to be deployed in dynamic environments where rapid responses are necessary, this latency must be reduced. Overcoming this limitation could involve exploring more efficient generative model architectures, model distillation techniques, or leveraging dedicated hardware acceleration.d for careful resource management in practical deployments.

\textbf{Social Impact.}
The approach presented in this paper stands to have a considerable impact on robotics research, industry, and society at large.

From a research perspective, our framework expands the horizon for robotic manipulation by demonstrating a viable path away from task-specific, data-intensive training. By leveraging foundation models, future research may require significantly fewer resources to develop capable robotic systems. This could accelerate the pace of innovation and contribute to the broader integration of intelligent robots in manufacturing, scientific exploration, and daily life.

Safety considerations are important to the responsible deployment of this technology. The generative model's decision-making process is not always transparent, posing a challenge for verification and validation. Therefore, rigorous monitoring and the implementation of robust safety protocols are essential, particularly in scenarios involving human-robot interaction or high-stakes operations. By emphasizing both the potential and the necessary precautions, this work encourages a responsible path forward in the advancement of cognitive robotics.

\section{Conclusion and Future Work}

In this work, we introduced an agentic, iterative framework for robotic manipulation, leveraging diffusion-based video models to generate executable trajectories from textual instructions. Our experimental evaluation demonstrated that the proposed agent significantly outperforms task-specific baselines across multiple manipulation tasks, with ANOVA confirming the effect of method on success rate ($F(4,60)=5.04, p=0.0014$). Humanoid platforms aligned with human-centric data, such as Unitree G1, showed the highest median success rates, though differences across embodiments were not statistically significant ($F(2,36)=2.01, p=0.1485$).

Physical robot trials further highlighted the benefit of iterative execution: both UR3 bimanual and G1 humanoid platforms achieved a final success rate of 80\%, substantially higher than first-attempt successes (30\% and 20\%, respectively). The majority of tasks required multiple corrective iterations, demonstrating the importance of the closed-loop reasoning–execution approach in recovering from initial failures.

Overall, our results indicate that diffusion-based video reasoning combined with iterative execution provides robust and generalizable performance across different platforms and tasks. Future work will focus on reducing computational overhead, enabling real-time applicability, and integrating safety-aware recovery strategies to handle irrecoverable failures, moving toward scalable, general-purpose robotic manipulation.

\bibliographystyle{IEEEtran}
\balance
\bibliography{bib}

\end{document}